\documentclass[preprint,showkeys,showpacs,preprintnumbers,amsmath,amssymb]{revtex4} 

\usepackage[american]{babel}

\usepackage[dvips]{graphicx}  
\usepackage{dcolumn}   
\usepackage{bm}        

\usepackage{amssymb, amsmath}
\usepackage{mathrsfs}			
\usepackage{color}

\newcommand{\highlight}[1]{\colorbox{yellow}{#1}}
\newcommand{\NA} {\backslash}

\newcommand{\mapping}{{\Phi}}		

 \newcommand{\ve}[1]{{\bf #1}}
 \newcommand{\vve}[1]{{\bf #1}}
 \newcommand{\veg}[1]{{\boldsymbol {#1}}}	
 \newcommand{\vveg}[1]{{\boldsymbol {#1}}}	

 \newcommand{\const} {{\rm const}}

\newcommand{\D}{\partial}

\newcommand{\dt}[1]{\frac {d #1} {d t}}
\newcommand{\delt}{{\Delta t}}

\newcommand{\DI}[1]{\frac {\D #1} {\D x_1}}	

\newcommand{\Di}[1]{\frac {\D #1} {\D x_i}}

\newcommand{\DIDI}[1]{\frac {\D^2 #1} {\D x_1^2}}

\newcommand{\DiDi}[1]{\frac {\D^2 #1 }{\D x_i^2 }}

\newcommand{\excl}[1]{{\backslash \hspace{-0.3em} #1}}

\newcommand{\crossout}[1]{{\slash \hspace{-0.5em} #1}}

\newcommand{\abs}[1]{\left|#1\right|}
\newcommand{\bracket}[1]{\left[#1\right]}

\newcommand{\parenth}[1]{\left(#1\right)}

\DeclareSymbolFont{AMSb}{U}{msb}{m}{n}
\DeclareMathSymbol{\R}{\mathbin}{AMSb}{"52}

\newcommand{\pf} {\noindent {\bf Proof.\ }}

\newtheorem{thm}{Theorem}[section]
\newtheorem{corollary}{Corollary}[section]
\newtheorem{defn}{Definition}[section]

\begin{document}

\title{
	Normalized multivariate time series causality analysis
	and causal graph reconstruction}

\author{X. San Liang}
\email{X.S. Liang, sanliang@courant.nyu.edu; http://www.ncoads.org/}
\affiliation{Nanjing Institute of Meteorology, Nanjing, China\\
    Shanghai Qizhi (Andrew C. Yao) Institute, Shanghai, China\\
	China Institute for Advanced Study,\\
    Central University of Finance and Economics, Beijing, China}



\begin{abstract}
Causality analysis is an important problem lying at the heart of science, 
and is of particular importance in data science and machine learning.
An endeavor during the past 16 years viewing causality as real physical
notion so as to formulate it from first principles, however, 
seems to go unnoticed.
This study introduces to the community this line of work, with a
long-due generalization of the information flow-based bivariate time series 
causal inference to multivariate series, based on the recent advance in 
theoretical development. 
The resulting formula is transparent, 
and can be implemented as a computationally very efficient 
algorithm for application.  
It can be normalized, and tested for statistical significance.
Different from the previous work along this line where only 
information flows are estimated, here an algorithm is also implemented
to quantify the influence of a unit to itself. While this forms a 
challenge in some causal inferences, here it comes naturally, and hence 
the identification of self-loops in a causal graph is fulfilled
automatically as the causalities along edges are inferred.

To demonstrate the power of the approach, 
presented here are two applications in extreme situations. The first is
a network of multivariate processes buried in heavy noises (with
the noise-to-signal ratio exceeding 100), 
and the second a network with nearly synchronized chaotic oscillators.
In both graphs, confounding processes exist.
While it seems to be a huge challenge to reconstruct from given series
these causal graphs, an easy application of the algorithm immediately
reveals the desideratum.
Particularly, the confounding processes have been accurately differentiated. 
Considering the surge of interest in the community, this study is very timely.
\end{abstract}


\maketitle

\section{Introduction}

Recent years have seen a surge of interest in causality analysis.
The main thrust is the recognition of its increasing importance 
in machine learning and artificial intelligence, a milestone being
the connection of the principle of independent causal mechanisms to
semi-supervised learning by Sch\"olkopf et al. (2012).
Different methods have been proposed for inferring 
the causality from data, in addition to the classical ones 
such as Granger causality testing. 
While traditionally causal inference has been categorized as a subject
in statistics, and now also a subject in computer science, 
it merits mentioning that, during the past decades, contributions
from different disciplines have augmented the subject 
and significant advances have been made ever since.
Early efforts since Clive Granger and Judea Pearl (c.f.~Pearl, 2009) 
include, for example, 
Spirtes and Glymour (1991),  Schreiber (2000), 
Palu$\check s$ et al. (2001), Liang and Kleeman (2005).
Recently, due to the rush in artificial intelligence, publications have
been piling up, among which are
Zhang and Spirtes (2008),
Maathuis et al. (2009),  
Pompe and Runge (2011),
Janzing et al. (2012),
Sugihara et al. (2012),
Sch\"olkopf et al. (2012),
Sun and Bollt (2014),
Peters et al. (2017),
to name but a few; 
see Peters et al. (2017) and Spirtes and Zhang (2016)
for more references.

%

Although causality has long been investigated ever since Granger (1969), 
thanks to the systematic works by Pearl (2009) and others, 
its ``mathematization is a relatively recent development,'' 
said Peters, Janzing and Sch\"okopf (2017).
On the other hand, Liang (2016) argued
that it is actually "a real physical notion 
that can be derived {\it ab initio}." 
Despite the current rush,
this latter line of work starting some 16 years ago, however, 
seems to go almost unnoticed. 
It can be traced back to a discovery on two-dimensional (2D)
information flow by Liang and Kleeman (2005).
With the efforts later on, e.g., Liang (2008) and Liang (2014), a very easy
method for bivariate time series causality analysis has been established,
validated, and applied successfully to real world problems in different
disciplines. More details can be found below in section~2.
Recently, the whole formalism has been put on a rigorous footing
(Liang, 2016); explicit formulas for multidimensional information flow
have been obtained in a closed form with both deterministic and stochastic
systems. 

The multivariate time series causality analysis, however, has not been 
established since Liang (2016)'s work. 
Considering the enormous interest in this field, we are henceforth intented
to fill the gap. 
The purpose of this study is hence two-fold:
(1) Implement Liang (2016)'s theory into the long-due multivariate time 
series causality analysis;
(2) along with the implementation present a brief 
introduction of this line of work.

The remaining of the paper is organized as follows. In section~2 we 
first establish the framework, and then take 
a stroll through the theory of information flow and the information flow-based 
bivariate time series causality analysis. Section~3 presents an estimate
of the information flow rates among multivariate time series, and their
significance tests. These information flows can be normalized to reveal the
impact of the role in question (section~4). In order to test the power of
the method, in section~5 it is applied to infer the causal graphs with 
two extreme processes, one being a network with heavy noise
(noise-to-signal ratio exceeding 100), another being a network of almost
synchronized chaotic oscillators. Section~6 closes the paper with a brief
summary of the study.

%

%
%

\section{An overview of the theory of information flow-based causality analysis}
\subsection{Directed graph, uncertainty propagation, and causality}

%
In this framework, causal inference is based on information flow (rather
than the other way around), which has been recognized as a real physical 
notion that can be put on a rigorous footing (see Liang, 2016). 
Consider a graph $(V,E)$, where $V$ and $E$
are the sets of vertexes and edges, 
and the structural causal model on the graph,
$({\mathcal C}, P_N)$, where $\mathcal C$ 
is a collection of $d$ structural assignments
$X_i = F_i({{\bf PA}(X_i)}, \epsilon_i)$, $i=1,...,d$,
${\bf PA}(X_i) \subseteq \{\ve X_{\excl i} \}
 = X_1, ..., X_d\} \backslash \{X_i\}$ 
indicating the parents or direct causes of $X_i$, and 
$P_N$ being a joint distribution over the noise variables
(Pearl, 2009). The basic idea is that this can be 
recast within the framework of dynamical systems, 
and that the causal inference problem can be carried forth to that
between the coordinates in a dynamical system. 
This is how Liang and Kleeman (2005) originally conceptualized the problem.
Recently it has also been realized by, say, R$\rm\crossout o$ysland (2012), 
Mooij et al. (2013) and Mogensen et al. (2018).

In physics there is a notion called information flow which can be readily
cast within the dynamical system framework. As entropy is by
interpretation ``self information'', it is natural to measure it with the
propagation of entropy or uncertainty, from one 
component to another. (Of course, other entropies may provide alternative
choices; e.g., Amig\'o et al., 2020)
In this light we have the following definition:
\begin{defn}
In a dynamical system $(\Omega, \mapping_t)$ on the $d$-dimensional
phase space $\Omega$,
where $\mapping_t$ may be a continuous-time flow ($t\in\R^+$) 
or discrete-time mapping $t\in\mathbb{Z}^+$),
the information flow from a component/coordinate $X_j$ 
to another component/coordinate $X_i$, written $T_{j\to i}$, 
is defined as the contribution of entropy (uncertainty) from $X_j$
per unit time ($t\in\R^+$) or per step ($t\in \mathbb{Z}^+$) 
in increasing the marginal entropy of $X_i$.
\end{defn}
%
%
With information flow, causality can be defined, and, moreover, 
quantitatively defined:
     \begin{defn}
     $X_j$ is causal to $X_i$ iff $T_{j\to i} \ne 0$. 
	The magnitude of the causality from $X_j$ to $X_i$
	is measured by $|T_{j\to i}|$. 
     \end{defn}
By evaluating the information flow within a dynamical system, the
underlying causal graph is henceforth determined.
For this study, we consider only the continuous flow case.
The vector field that form the structural assignments is hence
differentiable. Further assume a Wiener process for the noise (white noise).
Note that some of these assumptions can be easily relaxed, and 
the generalization is straightforward. But that is outside the scope of
this study.

\subsection{A brief stroll through the theory and recent advances}
This line of work begins with Liang and Kleeman (2005) within 
the framework of 2D deterministic systems. 
Originally it is based on a heuristic argument, but later on it is rigorized.
Its generalization to multidimensional and stochastic systems, however,
has not been fulfilled until the recent theoretical work by Liang (2016).
The following is just a brief review. 

We begin by stating an observational fact about causality:
	\begin{itemize}
	\item[$\ $]
   	{\it If the evolution of an event, say, $X_1$, is independent 
	of another one, $X_2$, then the information flow from $X_2$ to 
	$X_1$ is zero.}
	\end{itemize}
Since it is the only quantitatively stated fact about causality, 
all previous empirical/half-empirical causality formalisms have attempted 
to verify it in applications. For this reason, 
it has been referred to as the {principle of nil causality}
(e.g., Liang, 2016).
We will soon see below that, within the information flow framework, 
this principle turns out to be a proven theorem.

Consider a $d$-dimensional continuous-time stochastic system
for $\ve X = (X_1, ..., X_d)$
	\begin{eqnarray}	\label{eq:stoch_gov}
	d {\ve X} = \ve F(\ve X, t)dt + \vve B(\ve X, t) d{\ve W},
	\end{eqnarray}
where $\ve F = (F_1,..., F_d)$ 
may be arbitrary nonlinear functions of $\ve X$ and $t$,
${\ve W}$ is a vector of standard Wiener processes, and $\vve B  = (b_{ij})$ 
is the matrix of perturbation amplitudes 
which may also be any functions of $\ve X$ and $t$.
Assume that $\ve F$ and $\vve B$ are both differentiable with respect
to $\ve X$ and $t$. We then have the following theorem (Liang, 2016): 
%
%
	\begin{thm} 
	For the system (\ref{eq:stoch_gov}),
	the rate of information flowing from $X_j$ to $X_i$ (in nats per
	unit time) is
	\begin{eqnarray}	\label{eq:Tji}
	T_{j\to i} 
        &=& -E \bracket{\frac1{\rho_i} 
		       \int_{\R^{d-2}} \Di{(F_i\rho_{\excl j})} 
				d\ve x_{\excl i \excl j}} + \cr
	&&\qquad     \frac 12 E \bracket{\frac1{\rho_i} 
	    \int_{\R^{d-2}} \DiDi {(g_{ii}\rho_{\excl j})} 
				d\ve x_{\excl i \excl j}}, \cr
	&=&
	- \int_{\R^d} \rho_{j|i} (x_j|x_i) \Di {(F_i\rho_{\excl j})} d\ve x
	  +	\cr
	 &&\qquad \frac12 \int_{\R^d} \rho_{j|i} (x_j|x_i) 
		\DiDi {(g_{ii}\rho_\excl j)} d\ve x,
	\end{eqnarray}
	where $d\ve x_{\excl i \excl j}$ signifies 
	$dx_1 ... dx_{i-1} dx_{i+1} ... dx_{j-1} dx_{j+1}... dx_n$,
	$E$ stands for mathematical expectation, 
	$g_{ii} = \sum_{k=1}^n b_{ik} b_{ik}$, 
	$\rho_i = \rho_i(x_i)$ is the marginal probability density function
	(pdf) of $X_i$, $\rho_{j|i}$ is the pdf of $X_j$ conditioned on $X_i$,
	and $\rho_{\excl j} = \int_\R \rho(\ve x) dx_j$. 
	\end{thm}
For discrete-time mappings, the information flow is in a more
complicated form; see Liang (2016). 

%
%

\begin{corollary} 
When $d=2$,
\begin{eqnarray}	\label{eq:T21_2d}
T_{2\to1} = -E \bracket{\frac1{\rho_1} \DI{(F_1\rho_1)}}
	    + \frac12 E \bracket{\frac1{\rho_1} 
		\DIDI {g_{11}\rho_1}}.
\end{eqnarray}
\end{corollary}
This is the early result of Liang (2008) 
on which the bivariate causality analysis is based; see
Theorem~\ref{thm:L14} below.


There is a nice property for the above information flow: 
	\begin{thm} 
		\label{thm:PNC}
	If in (\ref{eq:stoch_gov})
	neither $F_1$ nor $g_{11}$ depends on $X_2$, 
	then $T_{2\to1} = 0$.
	\end{thm}
Note this is precisely the principle of nil causality. 
Remarkably, here it appears as a proven theorem, while
the classical ansatz-like formalisms attempt to verify in applications.


Moreover, Liang (2018) established that
	\begin{thm}  
	$T_{2\to1}$ is invariant under arbitrary nonlinear 
	transformation of $(X_3,X_4,...,X_d)$.
	\end{thm}
This is a very important result, as we will see soon in causal graph
reconstruction. On the other hand, this from an aspect tells that the
obtained information flow should be an intrinsic property in physical
world.

For linear systems, the information flow can be greatly simplified. 
	\begin{thm}
	In (\ref{eq:stoch_gov}), if $\ve F(\ve X) = \ve f + \vve A \ve X$, 
	and $\vve B$ is a constant matrix, then
	\begin{eqnarray}	\label{eq:Tji_linear}
	   T_{j\to i} = a_{ij} \frac {\sigma_{ij}} {\sigma_{ii}}, 
	\end{eqnarray}
	where $a_{ij}$ is the $(i,j)^{th}$ entry of $\vve A$,
	and $\sigma_{ij}$ the population covariance between
	$X_i$ and $X_j$.
	\end{thm}
Observe that, if $\sigma_{ij}=0$, then $T_{j\to i} = 0$; 
but if $T_{j\to i}=0$, $\sigma_{ij}$ does not necessarily vanish. 
Contrapositively, this means that correlation does not mean causation.
We hence have the following corollary:
	\begin{corollary}
	\begin{itemize}
	{In the linear sense
	causation implies correlation, but correlation does not
	 imply causation.}
	\end{itemize}
	\end{corollary}
This explicit mathematical expression hence provides a solution to 
the long-standing debate ever since George Berkeley (1710)
over correlation versus causation.

In the case with only two time series (no dynamical system is given), 
we have the following result (Liang, 2014): 
	\begin{thm} \label{thm:L14} 
	Given two time series $X_1$ and $X_2$, 
	under the assumption of a linear model with additive noise,
	the maximum likelihood estimator (mle) of (\ref{eq:T21_2d}) is
	\begin{eqnarray}	\label{eq:T21_est_2d}
	\hat T_{2\to1} = \frac {C_{11} C_{12} C_{2,d1} - C_{12}^2 C_{1,d1}}
			  {C_{11}^2 C_{22} - C_{11} C_{12}^2},
	\end{eqnarray}
	where $C_{ij}$ is the sample covariance
	between $X_i$ and $X_j$, 
	and $C_{i,dj}$ the sample covariance between $X_i$ and 
	a series derived from $X_j$ using the Euler forward differencing
	scheme:
	$\dot X_{j,n} = (X_{j,n+k} - X_{j,n}) / (k\delt)$, with $k\ge1$ 
	some integer.
	\end{thm}
Eq.~(\ref{eq:T21_est_2d}) is rather concise in form; 
it only involves the common statistics, i.e., sample covariances. 
In other words, a combination of some sample covariances will give 
a quantitative measure of the causality
between the time series. This makes causality analysis, which otherwise 
would be complicated with the classical empirical/half-empirical methods, 
very easy. Nonetheless, note that Eq.~(\ref{eq:T21_est_2d}) cannot replace
(\ref{eq:T21_2d}); it is just the mle of the latter. Statistical
significance test must be performed before a causal inference is made based
on the computed $T_{2\to1}$. For details, refer to Liang (2014).


The above formalism has been validated with many benchmark systems
such as baker transformation, H\'enon\ map, 
Kaplan-Yorke map, R\"ossler system 
(e.g., Liang (2016)), 
to name a few.
Particularly, the concise Eq.~(\ref{eq:T21_est_2d}) has been 
validated with problems where traditional approaches fail.
An example is the mysterious anticipatory system problem 
discovered by Hahs and Petehl (2011), 
which with (\ref{eq:T21_est_2d}) is successfully fixed in an easy way.



The formalism has been successfully applied to the studies of
many real world problems, among them are
the causal relation between El Ni\~no-Indian Ocean Dipole (Liang, 2014) 
global climate change (Stips et al., 2016), 
soil moisture-precipitation interaction (Hagan et al., 2018), 
glaciology (Vannitsem et al., 2019),
neuroscience problems (Hristopulos et al., 2019), 
to name a few. 
Here we particularly want to mention the study by Stips et al.\cite{Stips2016}
who, through examining with (\ref{eq:T21_est_2d}) the causality 
between the CO$_2$ index 
and the surface air temperature, identified a reversing causal relation
with time scale. They found, during the past century, indeed CO$_2$ emission
drives the recent global warming; the causal relation is
one-way, i.e., from CO$_2$ to global mean atmosphere temperature.
Moreover, they were able to find how the causality is distributed over
the globe, thanks to the quantitative nature of (\ref{eq:T21_est_2d}). 
However, on a time scale of 1000 years or over, 
the causality is completely reversed; that is to say, on a paleoclimate scale, 
it is global warming that drives CO$_2$ concentration to rise!

\section{Information flow among time series and algorithm for multivariate
causal inference}	\label{sect:estimate}

We now estimate (\ref{eq:Tji}), given observations of the $d$ components,
in order to arrive at a handy formula for causal inference.
As mentioned in section~1, this has not been done yet; the available
estimator (\ref{eq:T21_est_2d}) is for (\ref{eq:T21_2d}).
Here we only consider time series, but it can be easily 
extended to other forms of data.
We further assume the series are stationary and equi-distanced.
Without loss of generality, it suffices to examine $T_{2\to 1}$.

As in the bivariate case of Liang (2014),
we estimate the linear version (\ref{eq:Tji_linear}). We hence examine
a linear stochastic differential equation 
	\begin{eqnarray} 	\label{eq:gov}
	d\ve X = \ve f + \vve A \ve X dt + \vve B d\ve W,
	\end{eqnarray}
where $\ve f$ is a constant vector, and
$\vve A = (a_{ij})$ and 
$\vve B = (b_{ij})$ are constant matrices. Initially if $\ve X$ obeys a
Gaussian distribution, then it is a Gaussian for ever,
i.e., $\ve X \sim {\mathcal N} (\veg\mu,\vveg\Sigma)$,
with $\veg\mu=(\mu_1,...,\mu_d)^T$ and $\vveg\Sigma=(\sigma_{ij})$ 
being the mean vector and covariance matrix, respectively.
Hence $X_1 \sim {\mathcal N}(\mu_1, \sigma_{11})$.

The above results need to be estimated if what we are given are just 
$d$ time series.  That is to say, what we know is a single realization of some
unknown system, which, if known, can produce infinitely many realizations.
We use maximum likelihood estimation (e.g., Garthwaite et al., 1995) 
to achieve the goal. 
The procedure follows that of Liang (2014), 
which for easy reference we briefly summarize here.
As established before, a further assumption that $\vve B$ is diagonal,
i.e., $b_{ij}=0$, for $i\ne j$, and hence $g_{11}=b_{11}^2$, 
will much simplify the problem, while in practice this is
quite reasonable. 

Suppose that the series are equal-distanced with a time stepsize $\delt$,
and let $N$ be the sample size.
Consider an interval $[n\delt, (n+1)\delt]$, and let the transition pdf be
$\rho(\ve X_{n+1} | \ve X_n; \veg\theta)$, where $\veg\theta$ stands for
the vector of parameters to be estimated. So the log likelihood is
	\begin{eqnarray*}
	\ell_N(\veg\theta) =
	\sum_{n=1}^N \log\rho(\ve X_{n+1} | \ve X_n; \veg\theta)
	+ \log\rho(\ve X_1).
	\end{eqnarray*}
As $N$ is usually large, the term $\rho(\ve X_1)$ can be dropped without
causing much error. The
transition pdf is, with the Euler-Bernstein approximation 
(see Liang, 2014), 
	\begin{eqnarray*}
	&& \rho(\ve X_{n+1} = \ve x_{n+1} | \ve X_n = \ve x_n)
	= {[(2\pi)^d \det(\vve B \vve B^T \delt) ]^{-1/2}} \cr
	&&\quad
	  \times e^{-\frac12 (\ve x_{n+1} - \ve x_n - \ve F\delt)^T 
			     (\vve B\vve B^T\delt)^{-1} 
			     (\ve x_{n+1} - \ve x_n - \ve F\delt)},
	\end{eqnarray*}
where $\ve F = \ve f + \vve A\ve X$. This results in a log likelihood
functional
	\begin{eqnarray*}
	\ell_N(\ve f, \vve A, \vve B)
	= \const - \frac N2 \log \prod_i g_{ii} 
	 -\frac\delt 2 \parenth{
		\frac 1 {\sum_{i=1}^d g_{ii}} \sum_{n=1}^N R_{i,n}^2
				},
	\end{eqnarray*}
where 
	\begin{eqnarray*}
	R_{i,n} = \dot X_{i,n} - (f_i + \sum_{j=1}^d a_{ij} X_{j,n}),
	\qquad i=1,2,...,d
	\end{eqnarray*}
and $\dot X_{i} = \{\dot X_{i,n} \}$ 
is the Euler forward differencing approximation of $\dt {X_i}$: 
	\begin{eqnarray}	\label{eq:X_dot}
	\dot X_{i,n} = \frac {X_{i,n+k} - X_{i,n}} {k\delt},
	\end{eqnarray}
with $k\ge1$. Usually $k=1$ should be used to ensure accuracy, but in some
cases of deterministic chaos and the sampling is at the highest resolution,
one needs to choose $k=2$. Maximizing $\ell_N$,
it is easy to find 
that the maximizer 
	$(\hat f_1, \hat a_{11}, ... \hat a_{1d})$ 
satisfies the following algebraic equation:
	\begin{eqnarray}
	\bracket{\begin{array}{cccc}
          1                &{\overline{X_1}}   &... &{\overline{X_d}}   \\
	  {\overline{X_1}} &{\overline{X_1^2}} &... &{\overline{X_1X_d}}\\ 
	  \vdots	   &\vdots	       &\ddots &\vdots		\\
	  {\overline{X_d}} &{\overline{X_1X_d}}&... &{\overline{X_d^2}} \\ 
	\end{array}}
	\parenth{\begin{array}{c}
		\hat f_1	\\
		\hat a_{11}	\\
		\vdots		\\
		\hat a_{1d}	
	\end{array}}
	=
	\parenth{\begin{array}{c}
	\overline{\dot X_1}	\\
	\overline{X_1\dot X_1}	\\
	\vdots			\\
	\overline{X_d\dot X_1}
	\end{array}}
	\end{eqnarray}
where the overline signifies sample mean. 
After some algebraic manipulations as that in Liang (2014), 
this yields the maximum likelihood estimators (mle):
	\begin{eqnarray}
	&&\hat a_{1i} = \frac 1 {\det\vve C} \sum_{j=1}^d 
			\Delta_{ij} C_{j,d1} 	\label{eq:a1i} \\
	&&\hat g_{11} = \frac{Q_{N,1} \Delta t} N, \label{eq:g11} \\
 	&& \hat f_1 = \overline {\dot X_1} - \sum_{i=1}^d \hat a_{1i} \bar X_i,
	\end{eqnarray}
where 
	\begin{eqnarray}
	&&C_{ij} = \overline{(X_i - \bar X_i) (X_j - \bar X_j)},\\
	&&C_{i,dj} = \overline{(X_i - \bar X_i) (\dot X_j - \overline {\dot X_j})},
	\end{eqnarray}
are the sample covariances, $\Delta_{ij}$ the cofactors of the matrix 
$\vve C = (C_{ij})$, and
	\begin{eqnarray*}
	Q_{N,1} 
	&=& \sum_{n=1}^N \bracket{\dot X_{1,n} - 
	    (\hat f_1 + \sum_{j=1}^d \hat a_{1j} X_{j,n})}^2\cr
	&=& \sum_{n=1}^N \bracket{
	    (\dot X_{1,n} - \overline{\dot X_{1}})
	    - \sum_{i=1}^d \hat a_{1i} (X_{i,n} - \bar X_i) }^2  	\cr
	&=& N (C_{d1,d1} - 2\sum_{i=1}^d \hat a_{1i} C_{d1,i}
		 +  \sum_{i=1}^d \sum_{j=1}^d \hat a_{1i} \hat a_{1j} C_{ij}.
	\end{eqnarray*}

By (\ref{eq:Tji_linear}), 
this yields an estimator of the information flow from $X_2$ to $X_1$:
	\begin{eqnarray}	\label{eq:T21_est}
	\hat T_{2\to1} = \frac 1 {\det\vve C} \cdot 
		       \sum_{j=1}^d \Delta_{2j} C_{j,d1}
			\cdot \frac {C_{12}} {C_{11}},
	\end{eqnarray}
where $C_{j,d1}$ is the sample covariance between $X_j$ and the derived
series $\dot X_1$ as computed by (\ref{eq:X_dot}).
When $d=2$, it is easy to show that this is reduced to (\ref{eq:T21_est_2d}), 
the 2D estimator as obtained in Liang (2014).

Besides the estimator of information flow, in this study we actually have
also estimated the influence of a component on itself. 
	\begin{thm}
	Under a linear assumption, the maximum likelihood estimator 
	of $dH_1^*/dt$ is
	\begin{eqnarray}	\label{eq:T11_est}
	\widehat {\parenth{\dt {H_1^*} }} 
	= \frac 1 {\det\vve C} \cdot 
	       \sum_{j=1}^d \Delta_{1j} C_{j,d1}.
	\end{eqnarray}
	\end{thm}	
\pf
Since $dH_1^*/dt = E(\DI {F_1})$, which is $a_{11}$ in this case. The mle
hence follows. 

This supplies information not seen in previous causality analysis along
this line. As will be clear soon, this helps identify self loops in a causal
graph.

Statistical significance test can be performed for (\ref{eq:T21_est}) 
and (\ref{eq:T11_est}).
When $N$ is large, they are approximately normally distributed
around their true values with variances
$\parenth{\frac{C_{12}}{C_{11}}}^2 \hat \sigma^2_{a_{12}} $
and $\hat \sigma_{a_{11}}^2$, respectively,
thanks to the mle property. Here 
$\hat\sigma^2_{a_{12}}$
and $\hat\sigma^2_{a_{11}}$
are determined as follows (e.g., Garthwaite et al., 1995).  
Denote $\veg\theta = (f_1, a_{11}, a_{12},...,a_{1d}, b_1)$. Compute
	\begin{eqnarray*}
	I_{ij} = - \frac 1 N \sum_{n=1}^N 
		\frac{\D^2 \log\rho(\ve X_{n+1} | \ve X_n;\ \hat{\veg\theta})}
		     {\D\theta_i \D\theta_j}
	\end{eqnarray*}
to form a $(d+2)\times(d+2)$ matrix $\vve I$, 
namely, the Fisher information matrix.
The inverse $(N\vve I)^{-1}$ is the covariance matrix of $\hat{\veg\theta}$,
within which are $\hat\sigma_{a_{12}}^2$ and $\hat\sigma_{a_{11}}^2$. 
Given a significance level, the confidence interval can be found accordingly.

From the above an algorithm for causal inference hence can be implemented:
\begin{center}
\begin{tabular}{l}
\hline
\hline
{\bf Algorithm}: Quantitative causal inference\\
\hline
{\bf input:} $d$ time series\\
{\bf output:} a DG ${\mathcal G}=(V,E)$, and IFs along edges\\
	initialize $\mathcal G$ such that all vertexes are isolated; \\
	set a significance level $\alpha$; \\
{\bf foreach} $(i,j) \in V\times V$ {\bf do} \\
     \qquad compute $\hat T_{i\to j}$ by (\ref{eq:T21_est}); \\
     \qquad  {\bf if}  $\hat T_{i\to j}$ is significant at level $\alpha$  
							{\bf then} \\
 	\qquad\qquad add $i\to j$ to $\mathcal G$; \\
  	\qquad\qquad record $\hat T_{i\to j}$; \\
     \qquad {\bf end} \\
{\bf end} \\
return $\mathcal G$, together the IFs $\hat T_{i\to j}$\\
\hline
\end{tabular}
\end{center}

\section{Normalization of the causality among multivariate time series}

In many problems, just an assertion whether a causality exists is not
enough; we need to know how important it is. 
This raises an issue of normalization.
The normalization of information flow is by no means as trivial as it
seemingly looks. Quite different from the case as covariance vs.
correlation coefficient, no such relation as Cauchy-Schwartz inequality
exists. Liang (2015) listed some difficulties in the problem, and so far
this is still an arena of research. Hereafter we follow Liang (2015)
to propose the normalizer for  (\ref{eq:T21_est}).

The basic idea is that the normalizer for $T_{2\to1}$ should be related
to $dH_1/dt$, as the
former is by derivation a part of the contribution to the latter.
However, $dH_1/dt$ itself cannot be the normalizer, since many terms 
tend to cancel; sometimes $dH_1/dt$ may even completely vanish, 
just as in the H\'enon map case. 
We now write out the estimator of $dH_1/dt$ and see how 
the problem can be fixed.

   %

By Liang (2016), 
the time rate of change of the marginal entropy of $X_1$ is
	\begin{eqnarray}	\label{eq:dH1}
	\dt {H_1} = - E\parenth{F_1 \DI {\log\rho_1}} 
		    - \frac12 E\parenth{g_{11}\DIDI {\log\rho_1}}.
	\end{eqnarray}
In this linear case, 
	\begin{eqnarray}
	\dt {H_1} 
	&=& -E\parenth{\sum_{j=1}^d a_{1j} X_j \DI {\log\rho_1}}
		    - \frac12 E\parenth{g_{11} \DIDI {\log\rho_1} } \cr
	&=& E \parenth{\frac {X_1-\mu_1} {\sigma_{11}} \sum_j a_{1j} X_j}
	    + \frac 12 \frac {g_{11}} {\sigma_{11}}	\cr
	&=& a_{11} +  \sum_{j=2}^d T_{j\to1} + 
	      \frac 12 \frac {g_{11}} {\sigma_{11}}.
	\end{eqnarray}
The first term is $dH_1^*/dt$, i.e., the contribution from itself,
and the last term is the effect of noise, written $dH_1^{noise}/dt$. 
The remaining parts are the information flows to $X_1$, just as expected.
We may hence propose a normalizer as follows:
	\begin{eqnarray}
	Z \equiv  \abs{\dt {H_1^*}} + \sum_{j=2}^d \abs{T_{j\to1}}
		+ \abs{\dt {H_1^{noise}} }.
	\end{eqnarray}
And hence the normalized information flow from $X_2$ to $X_1$ is:
	\begin{eqnarray}
	\tau_{2\to1} = \frac {T_{2\to1}} Z.
	\end{eqnarray}
Clearly, $\tau_{2\to1}$ lies on $[-1,1]$. So, when $|\tau_{2\to1}|$ is 100\%, 
$X_2$ has the maximal impact on $X_1$.

Note that $\dt H_1^{noise} = g_{11} / (2\sigma_{11})$,
where 
	   $g_{11} = \sum_{j=1}^d b_{1j}^2$
is always positive. That is to say, noise always contributes
to increase the marginal entropy of $X_1$, agreeing with our common
sense. Obviously, this term is related to the noise-to-signal ratio.

By the results in section~{\ref{sect:estimate}}, $Z$ can be estimated as
	\begin{eqnarray}
	\hat Z = \abs{\widehat{ \parenth{\dt {H_1^*} } }} 
		+ \sum_{j=2}^d \abs{\hat T_{j\to1}}
		+ \abs{ \widehat{\parenth{\dt {H_1^{noise}}} }}.
	\end{eqnarray}
where
	$ \widehat{\parenth{\dt {H_1^{noise}}} } 
	  = \frac 12 \frac {\hat g_{11}} {C_{11}}$,
and $\hat g_{11}$,
$\widehat{\parenth{\dt {H_1^*}}}$ and $\hat T_{2\to1}$ are evaluated
using (\ref{eq:g11}), 
(\ref{eq:T11_est}) and (\ref{eq:T21_est}), respectively.

\section{Application to causal graph reconstruction}
\subsection{A noisy causal network from autoregressive processes}

\begin{figure}[h]
	\begin{center}
	\includegraphics[angle=0, width=0.5\textwidth, height=0.33\textwidth] 
				{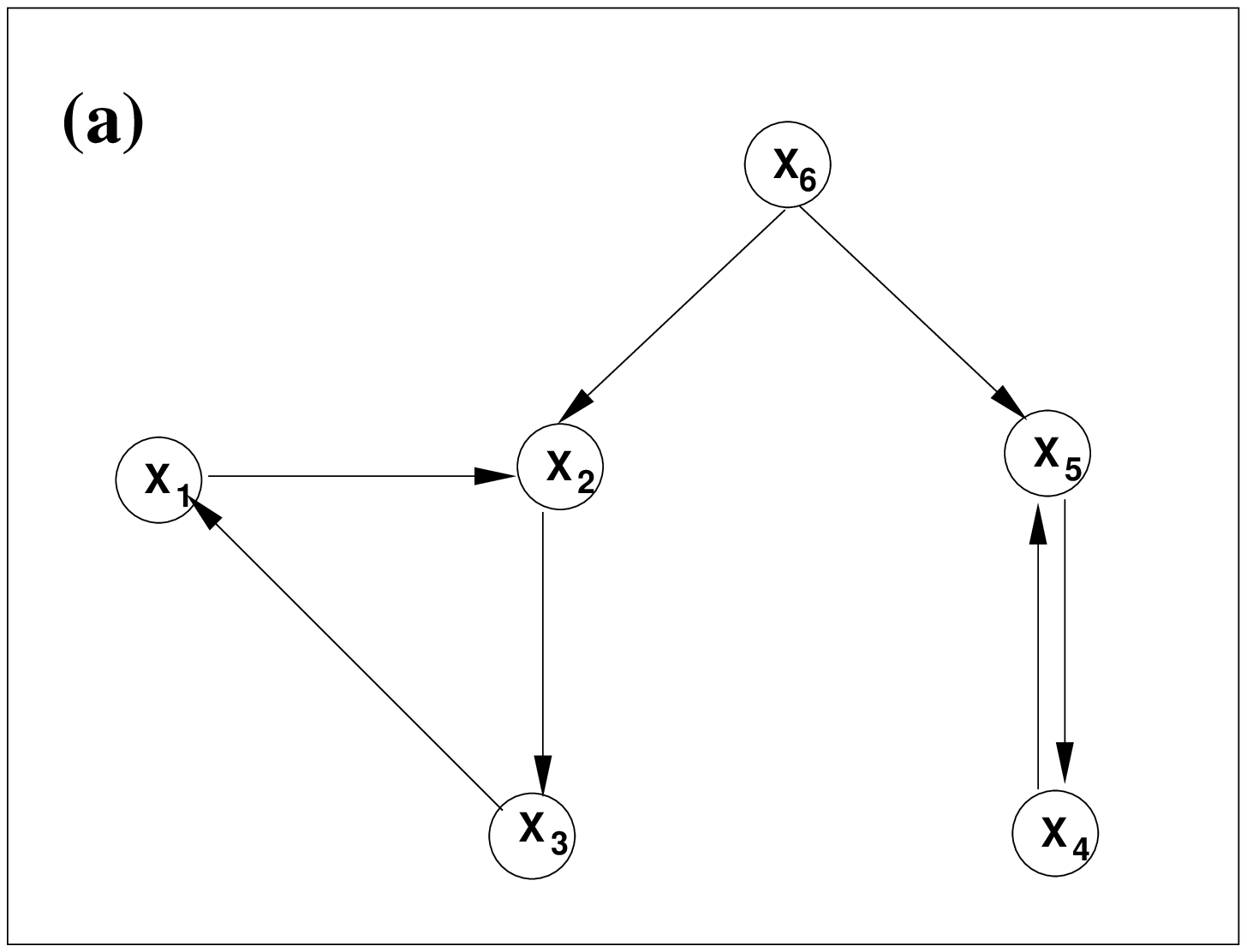}
	\includegraphics[angle=0, width=0.5\textwidth, height=0.32\textwidth] 
				{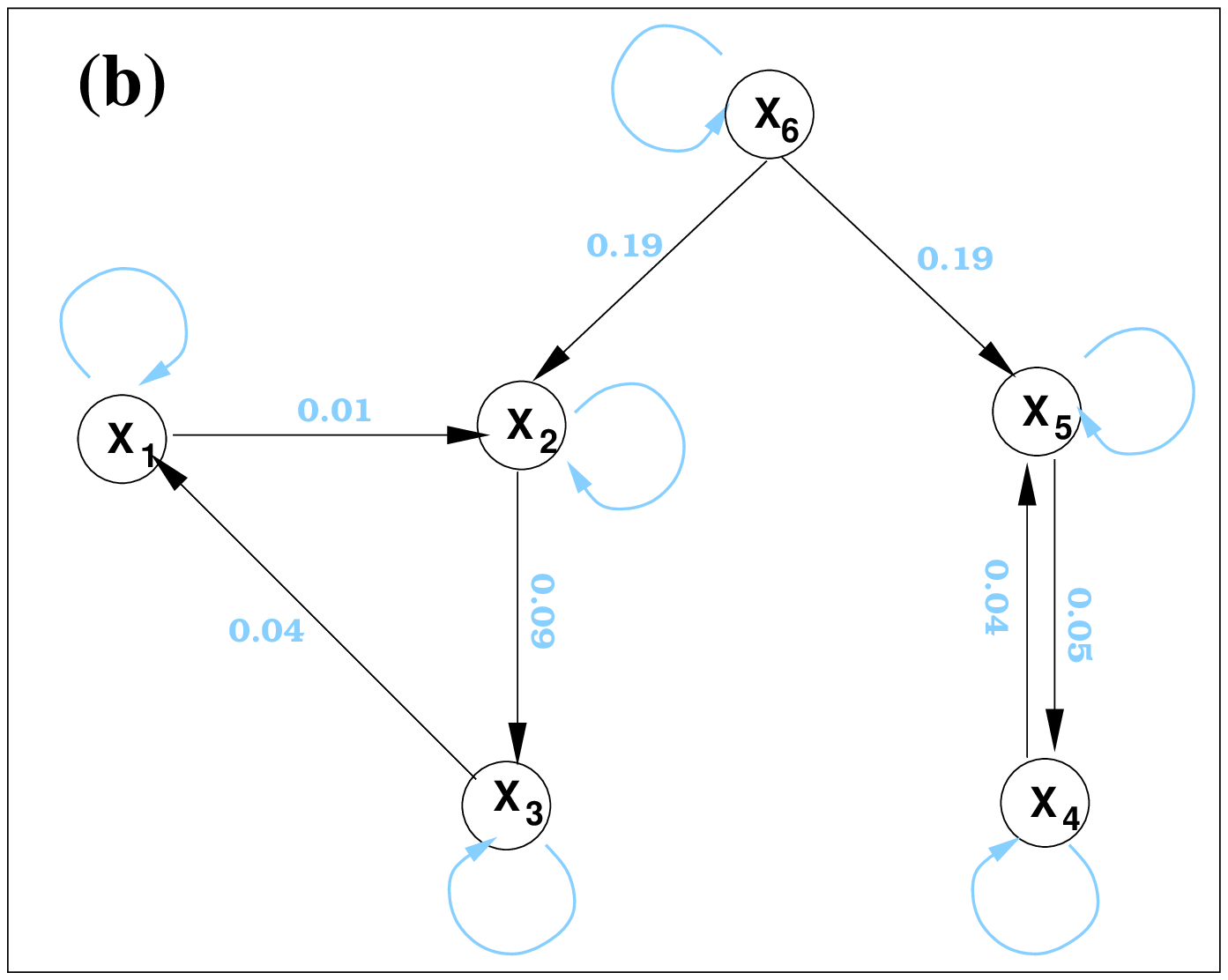}
	\caption{
	(a) A schematic of the directed network generated with the
	vector autoregressive processes (\ref{eq:var6}). 
	(b)
	The directed graph reconstructed from the six time series.
	Overlaid numbers
	are the respective significant information flows
	(in nats per time step); also overlaid are the inferred self loops
	or trivial cycles of length 1 (in light blue).
	} \label{fig:net6_values}
	\end{center}	
	\end{figure}

Consider the series generated from an $d$-dimensional 
vector autoregressive (VAR) process:
	\begin{eqnarray}	\label{eq:var6}
	\ve X(n+1) = \veg\alpha + \vve A \ve X(n) + \vve B \ve e(n+1)
	\end{eqnarray}
where $\ve X=(X_1,...,X_d)^T$, $\vve A = (a_{ij})_{d\times d}$,
$\ve e = (e_1,...,e_d)^T$,
$\vve B$ is a diagonal matrix with diagonal entries $b_{ii}$, $i=1,...,d$.
Here the errors $e_i \sim N(0,1)$ are independent, and $b_i$ are the
amplitudes of stochastic perturbation.
Let 
	\begin{eqnarray*}
	&& \vve A = \left(\begin{array}{cccccc}
		0  	&0	&-0.6	&0	&0	&0	\\
	       -0.5	&0  	&0	&0	&0	&0.8	\\
		0	&0.7	&0  	&0	&0	&0	\\
		0	&0  	&0	&0.7	&0.4	&0	\\
		0	&0	&0	&0.2	&0  	&0.7	\\
		0	&0	&0	&0	&0	&-0.5	
	\end{array}\right), 		\\
	&& \veg\alpha = (0.1, 0.7, 0.5, 0.2, 0.8, 0.3)^T,
	\end{eqnarray*}
The formed network is as shown in Fig.~\ref{fig:net6_values}a.
So by design we have two directed cycles ($X_1$,$X_2$,$X_3$) 
and ($X_4$,$X_5$).
The former is of length 3, while the latter are parallel edges.
These cycles are driven by a common cause or confounder $X_6$. 
Since no diagonal entries of $\vve A$ is 1,
all nodes are self loops (trivial cycles of length 1). 
The resulting autocorrelation is believed
to be a challenge in causal inferences for some techniques.
This and the confoundingness of $X_6$, 
have been two major issues for many causal inference methods.

First consider the case $b_{ii} = 1$.
Accordingly six series of 10000 steps are generated (randomly initialized). 
By computation the information flow rates are (only absolute values are
shown), 
if arranged in a matrix form such that the
$(i,j)^{\rm th}$ entry indicates $|T_{i\to j}|$,
then the absolute information flow rates are
	\begin{eqnarray*}
	\left(\begin{array}{cccccc}
	\NA   &\highlight{0.01}\    &0.00\    &0.00\   &0.00\   &0.00\\
	0.00\ &\NA      &\highlight{0.09}     &0.00\   &0.00    &0.00\\
	\highlight{0.05} &0.00      &\NA      &0.00    &0.00    &0.00\\
	0.00  &0.00     &0.00       &\NA  &\highlight{0.04}  &0.00\\
	0.00  &0.00     &0.00   &\highlight{0.05}       &\NA    &0.00\\
	0.00  &\highlight{0.19} &0.00  &0.00  &\highlight{0.18} &\NA
	\end{array}\right), 
	\end{eqnarray*}
So the only significant information flows are $T_{1\to2}$, $T_{2\to3}$,
$T_{3\to1}$, $T_{4\to5}$, $T_{5\to4}$, $T_{6\to2}$, $T_{6\to5}$, as
indicated in Fig.~\ref{fig:net6_values}b. (At a 90\% confidence level the maximal 
error is 0.005, so all these values are significant.)
This is precisely the same
as designed. So the causal graph is accurately reconstructed.
Also by (\ref{eq:T11_est}) 
$\abs{dH_1^*/dt}, ..., \abs{dH_6^*/dt}$ can be computed. They are:
$1.00\pm 0.01$, $1.01\pm0.01$, $1.01\pm0.01$, $0.30\pm0.01$, $1.00\pm0.01$,
$1.49\pm0.02$, where the errors at a 90\% confidence level are shown.
So here all the nodes are self loops (trivial cycles of length 1).

It should be particularly pointed out that
the confoundingness of $X_6$ does not make an issue here. 
As shown in Fig.~\ref{fig:net6_values}, 
there is no significant information flow between
$X_2$ and $X_5$; in other words, they are not directly causal to each other. 
Nor are $X_3$ and $X_4$. 
This is actually not a surprise; it is a corollary of the principle of nil
causality, as proved before (see Theorem~\ref{thm:PNC}).
Considering the difficulty of this problem,
the performance of this concise formula (\ref{eq:T21_est}) 
is remarkable.

The above information flows can be normalized to understand the impact of
one unit on another. For example, $|\tau_{6\to2}| = 13.2\%$,
$|\tau_{6\to5} = 12.5\%$.
For another example, in the cycle ($X_4$, $X_5$), the relative information
flows are $\tau_{4\to5} = 2.4\%$, $\tau_{5\to4} = 8.8\%$, in
contrast to the almost identical absolute information flows. This is
understandable: though $T_{5\to4}$ is comparable to $T_{4\to5}$, the parts
contributing to $dH_5/dt$ are different from that to $dH_4/dt$, 
and thus they may have different weights. 
 
Now consider an extreme case when the signals are buried within heavy
noise. Let $b_{ii} = 100$, and repeat the above steps. The results are,
remarkably, almost the same. 
So the formula (\ref{eq:T21_est}) is very robust in the presence of noise. 

If the time series is short, the performance is still satisfactory. For
example, if it has only 500 data points, the above case with heavy noise 
($b_{ii}=100$) results in the following matrix of information flow rates:
	\begin{eqnarray*}
	\left(\begin{array}{cccccc}
	\NA   &\highlight{0.02}\    &0.00\    &0.00\   &0.00\   &0.00\\
	0.00\ &\NA      &\highlight{0.13}     &0.00\   &0.01    &0.01\\
	\highlight{0.04} &0.01      &\NA      &0.00    &0.00    &0.00\\
	0.00  &0.00     &0.00       &\NA  &\highlight{0.07}  &0.00\\
	0.01  &0.00     &0.00   &\highlight{0.06}       &\NA    &0.00\\
	0.00  &\highlight{0.17} &0.00  &0.00  &\highlight{0.19} &\NA
	\end{array}\right), 
	\end{eqnarray*}
with the corresponding errors at the 90\% confidence level being:
	\begin{eqnarray*}
	\left(\begin{array}{cccccc}
	\NA   &0.00\ \    &0.00\ \    &0.00\ \   &0.00\ \    &0.01\\
	0.00\ \ &\NA      &{0.01}   &0.00\   &0.02    &0.02\\
	0.00  &0.01     &\NA      &0.00    &0.01    &0.01\\
	0.00  &0.00     &0.01     &\NA     &0.01    &0.00\\
	0.01  &0.00     &0.00     &{0.06}  &\NA     &0.02\\
	0.01  &0.01     &0.01     &0.00    &0.02    &\NA
	\end{array}\right).
	\end{eqnarray*}
So the significant (at the 90\% level) information flows are still those as highlighted.

\subsection{A network of nearly synchronized chaotic series}

Now consider the following causal graph made of R\"ossler oscillators $X$,
$Y$ and $Z$, where $X$ is a confounder. 
A R\"ossler oscillator has three components, so the system actually has a
dimension 9.

\begin{center}
\setlength{\unitlength} {1mm}
\begin{picture}(60,40)
   			\put(40,30){\circle*{10}}
   	         \put(38.25,28.5){\textcolor{white}{$X$}}
       \put(20,10){\circle*{10}}      		\put(60,10){\circle*{10}}
\put(18.25,8.5){\textcolor{white}{$Y$}}  \put(58.25,8.5){\textcolor{white}{$Z$}}

    \thicklines
    \put(36,26){\vector(-1,-1){12}}
    \put(44,26){\vector(1,-1){12}}
\end{picture}
\end{center}


We use for this purpose
the coupled system investigated by Palu$\rm\check s$ et al. (2018)
and Palu$\rm\check s$ and Vejmelka (2007). 
The 9 series are generated through the following R\"ossler systems
	\begin{eqnarray*}
	&&\left\{
	\begin{array}{l}
	d x_1 /dt = -\omega_1 x_2(t) - x_3(t),	\\
	 dx_2/dt = \omega_1 x_1(t) + 0.15 x_2(t),	\\
	 dx_3/dt = 0.2 + x_3(t) [x_1(t) - 10],
	\end{array}\right.	\\
	&&\left\{
	\begin{array}{l}
	 dy_1/dt = -\omega_2 y_2(t) - y_3(t) + 
		     \epsilon[x_1(t) - y_1(t)],	\\
	 dy_2/dt = \omega_2 y_1(t) + 0.15 y_2(t),	\\
	 dy_3/dt = 0.2 + y_3(t) [y_1(t) - 10],		
	\end{array}\right.	\\
%
	&&\left\{
	\begin{array}{l}
	 d {z_1}/dt = -\omega_3 z_2(t) - z_3(t) + 
		     \epsilon[x_1(t) - z_1(t)],	\\
	 d {z_2}/dt = \omega_3 z_1(t) + 0.15 z_2(t),	\\
	 d {z_3}/dt = 0.2 + z_3(t) [z_1(t) - 10].	
	\end{array}\right.	\\
	\end{eqnarray*}
Clearly, the first is the driving or ``master'' system, while the latter 
two are slaves which are not directly connected. We hence use them to define
$X$, $Y$ and $Z$.
This system is exactly the same as the one studied in 
Palu$\rm\check s$ et al. (2018), except for the addition of another
subsystem, $Z$. 
The parameters are also chosen the same as theirs: $\omega_1=1.015$ 
and $\omega_2=0.985$, but with an additional one: $\omega_3=0.95$.
As can be seen, $X$ is coupled with $Y$ and $Z$ through the first component, 
and the coupling is one-way, i.e., from $X$ to $Y$ and from $X$ to $Z$.
The coupling parameter $\epsilon$ is left open for tuning.

The above equations are differenced 
and the system is solved using the second order 
Runge-Kutta scheme with a time stepsize $\Delta t = 0.001$.
Initialized with random numbers, the state is integrated forward for
$N = 50000$ steps ($t=50$). Discard the initial 10000 steps and form
the 9 time series with 40000 data points.

\begin{figure}[h]
	\begin{center}
	\includegraphics[angle=0, width=0.75\textwidth] {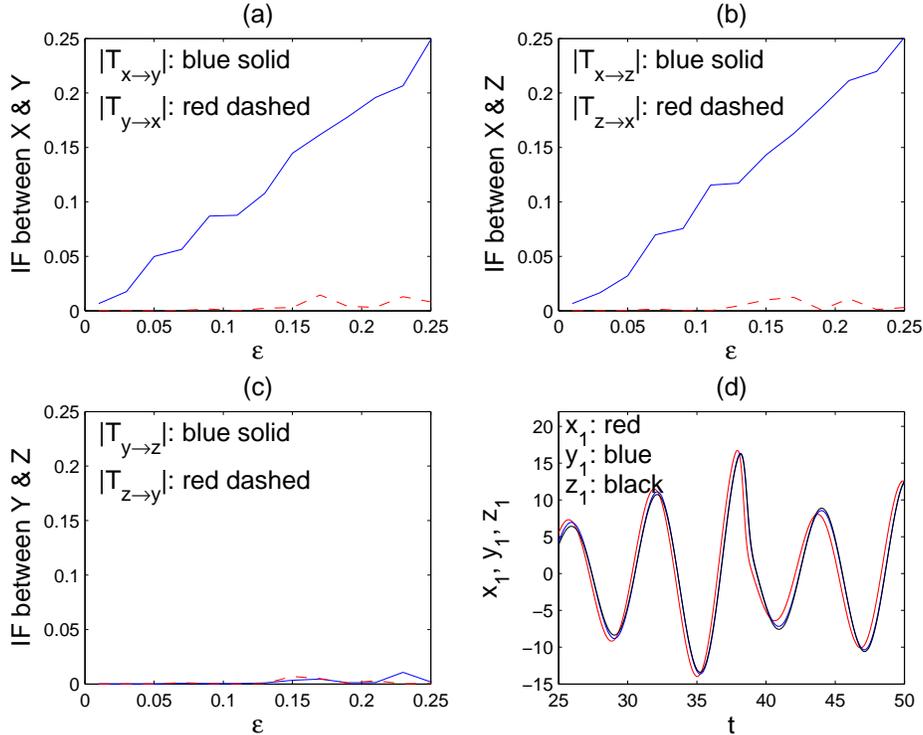}
	\caption{
	The information flows among the oscillators $X$, $Y$, and $Z$ 
	(in nats/unit time) versus the coupling strength $\epsilon$:
	(a) $|T_{X\to Y}|$ (blue) and $|T_{Y\to X}|$ (red);
	(b) $|T_{X\to Z}|$ (blue) and $|T_{Z\to X}|$ (red); 
	(c) $|T_{Y\to Z}|$ (blue) and $|T_{Y\to Z}|$ (red).
	(d) The series of $X_1$, $Y_1$, and $Z_1$ on a time interval 
	    when the coupling parameter $\epsilon=0.25$.
	} \label{fig:rossler}
	\end{center}	
	\end{figure}

The oscillators are highly chaotic. As $\epsilon$ increases, the three
oscillators gradually become in pace. They become almost synchronized after
$\epsilon>0.15$. Shown in Fig.~\ref{fig:rossler}d is an episode of the
synchronization for $\epsilon=0.25$.

We now apply (\ref{eq:T21_est}) to compute 
the information flows among $X$, $Y$, and $Z$. 
Since this is deterministic chaos problem, choose $k=2$ in 
(\ref{eq:X_dot}) and
(\ref{eq:T21_est}).
Following Palu$\rm\check s$ et al. (2018), the series $\{x_1(n)\}$, 
$\{y_1(n)\}$, and $\{z_1(n)\}$ are used to represent the three oscillators.
Shown in Figs.~\ref{fig:rossler}a, b, and c are dependencies of the
computed information flows on the coupling strength $\epsilon$. 
Clearly, among the six information flows, only $T_{X\to Y}$ and $T_{X\to Z}$
are significant, indicating (1) that the causal relation between $X$ and
$Y$ is unidirectionally from $X$ to $Y$, (2) that the causality between $X$ and
$Z$ is also one-way, i.e., from $X$ to $Z$, and, mostly importantly 
(3) that no direct causality exists between $Y$ and $Z$, although they are highly
correlated (c.f.~Fig.~\ref{fig:rossler}d). So here the confoundingness is
not at all an issue.


After $\epsilon$ exceeds 0.15, the systems begin to synchronize (see
Palu$\check s$ et al., 2018), and it is impossible to infer the causal
relation using traditional methods. 
This is understandable, as the series gradually approach to one series.
Here, however, even with $\epsilon>0.15$, i.e., even after the series are
almost synchronized, in this framework the inference still performs
remarkably well, as clearly seen in Figs.~\ref{fig:rossler}a, b, and c.
This attests to the power of the information flow-based causal inference
technique, which is however concise and very easy to implement.

\section{Conclusions}

Recent years have seen a surge of interest in causality analysis. 
This study introduced a line of work starting some 16 years ago which
however goes almost unnoticed, and implemented the state-of-the-art theory
(Liang, 2016) into a handy algorithm.
Particularly, this study extended the bivariate time series analysis 
of Liang (2014) to the long-due multivariate time series causal inference.

In a multivariate stochastic system, the information flow from one
component to another proves to be (\ref{eq:Tji}). 
When only time series are available, it can be estimated using
(\ref{eq:T21_est}) under a linear assumption. 
Ideally if it is not zero, then there exists
causality between the components, but practically statistical
significance need to be tested. These have been easily implemented as
an algorithm for use.

More than just finding the information flows, hence the causalities, among
the units, as did in Liang (2014), we have also estimated the influence of
a unit to itself. This results in the autocorrelation which becomes an
issue in some causal inferences. 
The consequence
is that, in a causal graph, those nodes which are self loops (cycles of
length 1) can be easily identified.
Also different from previous studies, in a unified treatment, 
the role of noise has been quantified along with the causality analysis.
This quantity has an easy physical interpretation, namely, the ratio of
noise to signal. Besides, the obtained causalities can be normalized to
measure the importance of the respective parental nodes.
It is shown that the normalizer should
be the sum of the absolute values of the Lyapunov exponent, the information
flows, and the noise contribution.

The above very concise and transparent 
formulas have been applied to examine two problems
in extreme situations: (1) a network of multivariate processes with heavy noise 
(stochastic perturbation amplitude 100 times the signal amplitude); 
(2) a network with nearly synchronized oscillators.
And, besides, confounding processes exist in both causal graphs.
Case (1) is made of vector autoregressive processes. By applying the
algorithm, the causal graph is accurately recovered in a very easy and
efficient way. 
Particularly, the confounding processes have be accurately clarified. 

In case (2), the network is formed with three chaotic R\"ossler oscillators. 
When the coupling coefficient exceeds a threshold, synchronization occurs.
However, even with the almost completely synchronized time series, the
information flow approach still performs remarkably well, with the
causalities accurately inferred, and the causal graph accurately
reconstructed. Particularly, the one-way causalities between the
master-slave systems have been recovered. 
Moreover, it is accurately shown that the two highly correlated, almost
identical series due to the confounder are not causally linked. 

It should be mentioned that, in arriving at the concise formula for causal
inference, an assumption of linearity has been invoked. For some nonlinear
problems the inference may not be precisely as expected. 
For example, in Figs.~\ref{fig:rossler}a and b, the red dashed lines are
supposed to be zero, but here they are not. But qualitatively the inference
is still good, as the one-way causality is clearly seen. 
Such success has already been evidenced in the bivariate case 
of Liang (2014), where a highly nonlinear problem defying classical 
approaches is examined. 
But, nonetheless, the power of the information flow-based causality
analysis won't be fully realized until the linear assumption is relaxed. 
To generalize to the fully nonlinear case is hence the goal of the next
step.




\begin{acknowledgements} 
This research is partially supported by National Science Foundation of
China under grant number 41975064.

\end{acknowledgements}

\end{document}